\documentclass[runningheads]{llncs}
\usepackage[T1]{fontenc}
\usepackage{graphicx}

\usepackage{hyperref}   % For hyperlinks in the PDF
\usepackage{cleveref}   % For clever references (import last)
\usepackage{multirow}   % For multirow in tables
\usepackage{booktabs}   % For better tables
\usepackage{caption}
\usepackage{subcaption} % For subfigures

\usepackage{color}
\usepackage{graphicx} 
\usepackage{xcolor,soul,colortbl}

\begin{document}
\title{Benchmarking Federated Machine Unlearning methods for Tabular Data}
\titlerunning{Benchmarking Federated Machine Unlearning}
% Chenguang Xiao (University of Birmingham) <x.a.trust@outlook.com>
% Abhirup Ghosh (University of Birmingham) <a.ghosh.1@bham.ac.uk>
% Han Wu (University of Southampton) <H.Wu@soton.ac.uk>
% Shuo Wang (University of Birmingham) <s.wang.2@bham.ac.uk>
% Diederick van Thiel (AdviceRobo) <diederick@advicerobo.com>
\author{
    Chenguang Xiao\inst{1} \and
    Abhirup Ghosh\inst{1} \and
    Han Wu\inst{2} \and
    Shuo Wang\inst{1} \and
    Diederick van Thiel\inst{3}
}

\authorrunning{C. Xiao et al.}
\institute{
    University of Birmingham, Birmingham B15 2TT, UK \and
    University of Southampton, Southampton SO17 1BJ, UK \and
    AdviceRobo, Den Haag 2516 AH, Netherlands
}
\maketitle              % typeset the header of the contribution
\begin{abstract}
    % by Diederick
    Machine unlearning, which enables a model to forget specific data upon request, is increasingly relevant in the era of privacy-centric machine learning, particularly within federated learning (FL) environments. This paper presents a pioneering study on benchmarking machine unlearning methods within a federated setting for tabular data, addressing the unique challenges posed by cross-silo FL where data privacy and communication efficiency are paramount. We explore unlearning at the feature and instance levels, employing both machine learning, random forest and logistic regression models. Our methodology benchmarks various unlearning algorithms, including fine-tuning and gradient-based approaches, across multiple datasets, with metrics focused on fidelity, certifiability, and computational efficiency. Experiments demonstrate that while fidelity remains high across methods, tree-based models excel in certifiability, ensuring exact unlearning, whereas gradient-based methods show improved computational efficiency. This study provides critical insights into the design and selection of unlearning algorithms tailored to the FL environment, offering a foundation for further research in privacy-preserving machine learning.
\end{abstract}
%
%
%

% \begin{table}[htbp]
%     \centering
%     \begin{tabular}{lll}
%         Section            & Author                   & Status      \\
%         \midrule
%         Introduction       & Shuo, Han, Abhirup       & In progress \\
%         Preliminaries      & Han, Diederick           & In progress \\
%         Related Work       & Han, Chenguang           & In progress \\
%         Data / Application & Shuo, Abhirup, Diederick & In progress \\
%         Methodology        & Chenguang                & Done        \\
%         Experiments        & Chenguang                & Done        \\
%         Results            & Chenguang                & Done        \\
%         Discussion         & All                      & In progress \\
%         \bottomrule
%     \end{tabular}
%     \caption{Sections and contributors.}
%     \label{tab:sections}
% \end{table}

\section{Introduction}
\label{intro}

Learning patterns from data while preserving user privacy is critical when finance or health organizations learn from their user pool. For example, think of predicting risk of giving credit to a customer or health diagnostic prediction from physiological markers. While such predictors naturally need a large pool of user data which necessitate collaboration among organizations, curating such a big data is impossible as user data is private and cannot cross legal boundaries.

The above applications primarily require two capabilities: \(i)\) training a machine learning model without sharing the data, and \(ii)\) flexibility to be able to remove the data from a trained model. They enable the participants to always control their data before, during, and after the training. Such requirement is inline with the legal regulations including General Data Protection Regulation (GDPR) \cite{gdprinfoGeneralData}, the California Consumer Privacy Act (CCPA) \cite{caCaliforniaConsumer}, the Act on the Protection of Personal Information (APPI) \cite{japaneselawtranslationProtectionPersonal}, and Canada’s proposed Consumer Privacy Protection Act (CPPA)~\cite{didomiCanadasData}.

Recently, two subdomains of machine learning have separately addressed the above two challenges \(i)\) FL enabling many participants (called as clients; E.g., finance organizations) collaboratively train a model under the orchestration of a central server (e.g., service provider), while keeping the data local in their individual sources~\cite{kairouz2021advances}; and \(ii)\) Machine unlearning~\cite{zhangReviewMachineUnlearning2023}, a reverse process of machine `learning' that forgets the information it has learned.

% FL is a machine learning setting where many clients (e.g., mobile devices or whole organizations) collaboratively train a model under the orchestration of a central server (e.g., service provider), while keeping the data local in their individual sources~\cite{kairouz2021advances}. 

% It is a collaboratively decentralized privacy-preserving technology to overcome challenges of data silos and data sensibility~\cite{li2020review}. This offers ample opportunities in critical domains such as healthcare, finance etc, where it is risky to share private user information to other organisations or devices \cite{Mammen2021fedlearn}.

% ulearning and FL
% Machine unlearning is the reverse process of machine learning, where the model is trained to forget the information it has learned.
% This is particularly important in the context of privacy and data protection consider the power of machine learning models to memorize the training data.
% FL is a distributed machine learning approach where the model is trained on multiple devices or servers holding the data.
% Raw data is invisible to the FL server by design for the privacy concerns during the training process.
% However, this does not guarantee the particularly sensitive data is not memorized by the model.
% In this setting, the model has to forget the information it has learned from the data on each device or server.

Both Federated Learning and machine unlearning are challenging problems. Federated learning iteratively aggregates partially trained models on clients using their local data. The task becomes non-trivial when the pattern across the clients differ significantly~\cite{bourtoule2021machine}. Especially due to the stochastic nature of deep learning models, there is no efficient way to mathematically remove data information from model parameters~\cite{nguyen2022survey}. Simply removing the corresponding data from the database is insufficient as the models already trained on the data retain critical information, which is especially true for today's large models that are prone to memorize training data~\cite{ullah2021machine}. Attackers are able to reconstruct or infer the private information in training data due to these memorization capability~\cite{rigaki2023survey,jegorova2022survey,wu2022federated}. This makes the problem of machine unlearning non-trivial.

While both federated learning and machine unlearning individually have attracted significant attention in the literature (Section ~\ref{related}), their combination is still under-explored. However, the setting is natural, for example, think of a FL system with a consortium of banks to learn a credit risk predictor without compromising privacy of their customers. With machine unlearning the banks can now enable their customers to withdraw their data contribution such as at the end of their contract.

The combined problem brings in additional challenges in terms of the wide variety of possible settings. These problems becomes natural in finance and health domain, especially with tabular data. To our best of knowledge, this is the first work to empirically explore a wide variety of such settings from an exhaustive set of performance metrics. In the following we brief the aspects we explore in the paper.

% as a majority of unlearning methods need access to the data to be forgotten~\cite{} which is impossible in federated setting as the data is never exposed to the server.

% Advice from Diederick
% On the relevance of privacy and data protection (security),  I would add 'regulatory compliance'
% Reference the GDPR European Commission , 2024) https://commission.europa.eu/law/law-topic/data-protection_en
% the European AI-act (European Commission, 2024) https://commission.europa.eu/news/ai-act-enters-force-2024-08-01_en
% DORA (digital operational resilience act) (European Commission, 2024) https://finance.ec.europa.eu/regulation-and-supervision/financial-services-legislation/implementing-and-delegated-acts/digital-operational-resilience-regulation_en
% UK GDPR (ICO, 2024) https://ico.org.uk/for-organisations/data-protection-and-the-eu/data-protection-and-the-eu-in-detail/the-uk-gdpr/
% California Privacy Act ( State of California, 2024) https://oag.ca.gov/privacy/ccpa

% unlearning
In machine unlearning, we focus on two aspects: Target of unlearning and method of unlearning.
In terms of assessing the unlearning methods, we consider the following metrics~\cite{zhangReviewMachineUnlearning2023}: Fidelity, Certifiability, and Efficiency.

% unlearning in FL
Machine unlearning in the FL system confronts the challenges from the unlearning algorithms with constraints from the FL regulations.
Generally, two FL regulations are considered in the unlearning process:
\(1)\) Data privacy: the FL server cannot access the data on the devices or servers.
\(2)\) Communication efficiency: the FL server cannot access the model on the devices or servers.
% There are two types of unlearning algorithms in the FL system~\cite{zhangReviewMachineUnlearning2023}:
% \(1)\) Exact unlearning: the model is same as the model trained without the data to be forgotten.
% \(2)\) Approximate unlearning: the model is similar to the model trained without the data to be forgotten.

\section{Related Work}
\label{related}

\subsection{Federated Learning}
FL is a decentralized learning paradigm where models are trained collaboratively across multiple devices or institutions without centralizing data.
Proposed by McMahan et al., FL allows models to be trained on distributed data, thus addressing concerns related to data privacy and communication costs~\cite{mcmahanCommunicationEfficientLearningDeep2017}.
Several enhancements have since been proposed to tackle issues such as communication efficiency, model heterogeneity, and security threats like poisoning attacks and adversarial examples~\cite{bonawitz2019towards,kairouz2021advances}.
However, these solutions primarily focus on learning and model optimization, with little attention to post-hoc model adjustments, such as unlearning.

\subsection{Machine Unlearning}
Machine unlearning refers to the process of efficiently removing specific training data points from a machine learning model without needing to retrain the model from scratch.
The foundational work by Cao and Yang formalized the unlearning problem and proposed techniques for probabilistically \textbf{forgetting} data~\cite{cao2015towards}.
This approach is motivated by privacy regulations such as \hyperlink{https://ico.org.uk/for-organisations/data-protection-and-the-eu/data-protection-and-the-eu-in-detail/the-uk-gdpr/}{GDPR}, which empower users with the right to request the deletion of their personal data.
Subsequent efforts~\cite{ginart2019making,golatkar2020eternal} explored efficient mechanisms for unlearning in specific models, such as support vector machines (SVMs) and deep neural networks.
While these methods significantly reduce the computational cost of retraining, most existing work assumes centralized data and model training, which does not align with FL's decentralized nature. Addressing these limitations, Bourtoule et al~\cite{bourtoule2021machine} introduced a technique known as 'sharding,' which involves dividing the data into independent partitions and constructing the final model from submodels trained on these shards. This approach allows for efficient unlearning by merely retraining affected submodels, enhancing the process for models like least-squares regression.

\subsection{Federated Machine Unlearning}
The intersection of FL and machine unlearning has gained increasing attention due to the rising demand for privacy-preserving machine learning.
Federated machine unlearning (FMU) is a nascent area where the goal is to remove specific data points or participants' contributions from a global model in a federated setting.

Initial approaches to FMU leverage differential privacy~\cite{abadi2016deep} to ensure that individual data points have minimal influence on the global model.
However, these techniques primarily offer theoretical guarantees and are often inefficient when the need for unlearning arises frequently or at large scales.
More recent proposals focus on model partitioning, federated pruning, and modular updates, where smaller portions of the model can be retrained or discarded based on the data being unlearned~\cite{wu2022federated}.

\section{Methodology}
\label{benchmark}
The methodology of unlearning used in this works is based on the scenarios of unlearning and the models to be unlearned.
According to the different scenarios of unlearning, we applied different unlearning algorithms to the models.

The model used for the unlearning task in this work are neural networks.
Neural networks are widely used in the FL system for the FL task with ability to learn complex patterns in the data.
Specifically, we use the logistic regression model as the neural network model and the random forest model as the decision tree model.
In FL setting, FedAvg~\cite{mcmahanCommunicationEfficientLearningDeep2017} is used as the base algorithm to aggregate the models from the clients.

\subsection{Scenarios of Unlearning}
% dimension of unlearning
% 1. unlearn given columns (features)
% 2. unlearn given rows (instances)
Two kinds of unlearning scenarios are considered regarding the dimension of data to be forgotten: unlearn given columns (features) and unlearn given rows (instances).
Unlearning given columns is to forget sensitive features in the data, e.g., the age and gender of the individuals.
Unlearning given rows is to forget certain data instances, e.g., the data from an individual who has withdrawn the consent.

For the unlearning of features, we set value of those features to be zero to consent the removal.
Setting them to zero is equivalent to removing them from the data, while keeping the data structure unchanged.
This is particularly useful as the unlearning process will not change the model architecture.
As all the rows are affected by the unlearning of features, there is no difference between the forgotten set and retained set in such case.

For the unlearning of instances, we remove the rows affected.
The set of instances to be forgotten is called the forgotten set, while the reset is called the retained set.

\subsection{Unlearning Algorithms}
% unlearning for 4 scenarios
The baseline algorithm for all the unlearning tasks is to \textbf{retrain} the model without the data to be forgotten.

For the unlearning of features with neural networks, we use a \textbf{fine-tuning} algorithm to update the model without the features to be forgotten.
For the unlearning of instances, three gradient-based unlearning algorithms are used to update the model without the instances to be forgotten.
They are the \textbf{gradient descent} method, the \textbf{gradient difference} method, and the \textbf{KL Minimization} method~/cite{mainiTOFUTaskFictitious2024}.

\section{Experiments}
\label{experiments}
In this section, we describe the experiments conducted to evaluate the unlearning algorithms in the FL setting.
In details, we describe the datasets used, the settings of the experiments, and the metrics used to evaluate the unlearning algorithms.

\subsection{Datasets}

5 public datasets and 1 private finance datasets are used in the experiments.
They are:
\(1)\) Bank Personal Loan Modelling (BL),
\(2)\) Diabetes (DB),
\(3)\) German Credit Data (GC),
\(4)\) Adult (AD),
\(5)\) Hospital Readmission Dataset (HD),
and \(6)\) Private (PR) datasets.

\textbf{Bank Personal Loan Modelling} dataset contains demographic information of 5000 customers.
Among these 5000 customers, only 480 (= 9.6\%) accepted the personal loan that was offered to them.
\textbf{Diabetes} dataset is originally from the National Institute of Diabetes and Digestive and Kidney Diseases.
There are 8 features and a label variable indicating whether the patient has diabetes.
\textbf{German Credit} dataset contains 1000 entries with 20 categorial/symbolic attributes prepared to infer the credit risk of the individuals.
\textbf{Adult} dataset contain information for predicting whether annual income of an individual exceeds \$50K/yr based on census data.
\textbf{Hospital Readmission} dataset represents ten years of clinical care data in US hospitals. It includes over 50 features which can be used for predicting hospital readmission.
\textbf{Private} dataset is our finance dataset with 18 features and a label variable indicating whether the customer's loan application will be approved.

When unlearning the features, we're unlearning 1 features in the datasets.
For each datasets, we chose 3-5 features to be unlearned.
When unlearning the instances, we're unlearning 5\% of the instances in the datasets.
The forgotten set in this case are randomly selected from the datasets.

\subsection{Metrics}
There are three factors we concern in the evaluation of the unlearning algorithms:
\(1)\) Fidelity: the performance of the model after unlearning.
\(2)\) Certifiability: the effectiveness of erasing the data from the model.
\(3)\) Efficiency: the computational cost of the unlearning process.

Given class imbalance are common in all the selected datasets as well as the real-world financial data, instead of overall accuracy, we use the F1 score as the metric for the fidelity.
We also consider the recall, and precision additionally to further show the bias of the unlearning algorithms.
We use F1, PPV, and TPR for short in the following sections to represent the F1 score, precision, and recall, respectively.

There are multiple ways to measure the certifiability of the unlearning algorithms.
The key idea of certifiability is to measure how well the unlearning algorithms forget the data.
As knowledge hidden in the model is hard to measure, existing certifiability metrics are based on the model's prediction on the forgotten set.
The baseline for the output is the retrained model without the forgotten set.
Finally, the similarity between the output and the baseline is used as the certifiability metric.
There are multiple ways to measure the similarity, e.g., KL-divergence, cosine similarity, etc.
Here in this work, we simply use the \textbf{residual norm} between the unlearning model output and the retrain output as the certifiability metric.

For the efficiency, we measure the iterations of the unlearning process.
In the FL setting, the communication cost is main concern in the efficiency which directly related to the global rounds in the FL system.

\section{Results and Analysis}
\label{results}
The experiments results are grouped into two part based on the dimension of the data to be forgotten: unlearn given columns (features) and unlearn given rows (instances).
Every result in the following section is the average with the standard deviation in brackets over 10 independent runs.

\subsection{Unlearn Columns}

\Cref{tab:fidelity} shows the fidelity of the unlearning algorithms on the datasets.
In these experiments, the first feature of each dataset is unlearned.
Generally, fine-tuning ranks the best in half of the metrics on six datasets, indicating the effectiveness of fine-tuning in unlearning features.
Considering the F1 score, retraining only show slightly better performance than fine-tuning on the Hospital Readmission Dataset at 5\%.
\begin{table}[htbp]
    \centering
    \caption{Fidelity of the unlearning algorithms on the datasets.}
    \label{tab:fidelity}
    \renewcommand{\arraystretch}{0.9}
    \begin{tabular}{llrrr}
        \toprule
        Data                   & Met & finetune                & retrain                 & train                   \\
        \midrule
        \multirow[c]{3}{*}{AD} & F1  & 0.551 (0.010)           & 0.589 (0.011)           & \bfseries 0.600 (0.010) \\
                               & TPR & \bfseries 0.910 (0.008) & 0.736 (0.033)           & 0.767 (0.012)           \\
                               & PPV & 0.395 (0.010)           & \bfseries 0.492 (0.018) & 0.492 (0.014)           \\
        \hline
        \multirow[c]{3}{*}{BL} & F1  & \bfseries 0.545 (0.071) & 0.495 (0.039)           & 0.456 (0.037)           \\
                               & TPR & 0.852 (0.071)           & 0.840 (0.067)           & \bfseries 0.898 (0.072) \\
                               & PPV & \bfseries 0.408 (0.084) & 0.353 (0.041)           & 0.309 (0.041)           \\
        \hline
        \multirow[c]{3}{*}{DB} & F1  & \bfseries 0.594 (0.040) & 0.585 (0.068)           & 0.588 (0.049)           \\
                               & TPR & \bfseries 0.923 (0.062) & 0.899 (0.119)           & 0.872 (0.091)           \\
                               & PPV & 0.441 (0.049)           & 0.439 (0.066)           & \bfseries 0.446 (0.046) \\
        \hline
        \multirow[c]{3}{*}{GC} & F1  & 0.550 (0.154)           & 0.521 (0.241)           & \bfseries 0.597 (0.219) \\
                               & TPR & 0.442 (0.209)           & 0.457 (0.281)           & \bfseries 0.561 (0.270) \\
                               & PPV & \bfseries 0.857 (0.081) & 0.797 (0.115)           & 0.755 (0.056)           \\
        \hline
        \multirow[c]{3}{*}{HD} & F1  & 0.565 (0.010)           & \bfseries 0.592 (0.025) & 0.588 (0.010)           \\
                               & TPR & 0.524 (0.021)           & \bfseries 0.599 (0.075) & 0.578 (0.023)           \\
                               & PPV & \bfseries 0.614 (0.010) & 0.592 (0.022)           & 0.598 (0.009)           \\
        \hline
        \multirow[c]{3}{*}{PR} & F1  & \bfseries 0.493 (0.105) & 0.465 (0.100)           & 0.471 (0.100)           \\
                               & TPR & \bfseries 0.492 (0.098) & 0.477 (0.103)           & 0.471 (0.087)           \\
                               & PPV & \bfseries 0.539 (0.083) & 0.524 (0.106)           & 0.518 (0.065)           \\
        \bottomrule
    \end{tabular}
\end{table}

The certifiability of the finetune method is shown in \Cref{tab:certifiability}.
The metric used is the residual norm between the unlearning model output and the retrain output.
The results show the mean and STD in brackets of the residual norm on the datasets.
\begin{table}[htbp]
    \centering
    \caption{Certifiability of the \textbf{finetune} on selected datasets.}
    \label{tab:certifiability}
    \begin{tabular}{rrrrrr}
        \toprule
        AD             & BL             & DB             & GC             & HD             & PR            \\
        \midrule
        0.148  (0.046) & 0.108  (0.037) & 0.220  (0.068) & 0.239  (0.071) & 0.208  (0.065) & 0.128 (0.040) \\
        \bottomrule
    \end{tabular}
\end{table}

\begin{figure}[htbp]
    \centering
    \begin{subfigure}[b]{0.49\linewidth}
        \centering
        \includegraphics[width=0.8\linewidth]{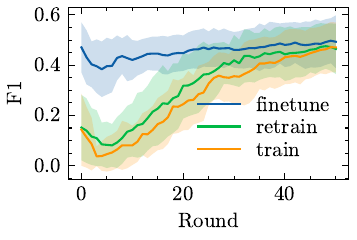}
        \caption{Unlearn feature of Private dataset.}
        \label{fig:dynata_f0_f1}
    \end{subfigure}
    \hfill
    \begin{subfigure}[b]{0.49\linewidth}
        \centering
        \includegraphics[width=0.8\linewidth]{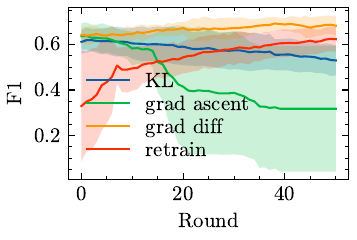}
        \caption{Unlearn rows of the Diabetes dataset.}
        \label{fig:row_DB_0.05_F1}
    \end{subfigure}
    \caption{F1 curve of the unlearning algorithms.}
\end{figure}

\Cref{fig:dynata_f0_f1} shows the efficiency of the unlearning algorithms on the datasets.
It is clear that the fine-tuning algorithm is more efficient one compared with the retraining.
There is a sharp decrease in the F1 score of the fine-tuning algorithm in the first few rounds, and then it remains stable.
The F1 curve of retraining follows the same pattern as the original training process.

\begin{table}[htbp]
    \centering
    \caption{Fidelity of the algorithms for the row unlearning task.}
    \label{tab:row_fidelity}
    \renewcommand{\arraystretch}{0.9}
    \begin{tabular}{llrrrr}
        \toprule
        Data                   & loss & KL                      & grad ascent             & grad diff               & retrain                 \\
        \midrule
        \multirow[c]{3}{*}{AD} & F1   & 0.581 (0.012)           & 0.153 (0.198)           & \bfseries 0.595 (0.010) & 0.592 (0.012)           \\
                               & TPR  & 0.758 (0.040)           & 0.400 (0.516)           & 0.768 (0.011)           & \bfseries 0.774 (0.011) \\
                               & PPV  & 0.472 (0.022)           & 0.095 (0.123)           & \bfseries 0.486 (0.012) & 0.479 (0.014)           \\
        \hline
        \multirow[c]{3}{*}{DB} & F1   & 0.530 (0.066)           & 0.318 (0.276)           & \bfseries 0.682 (0.042) & 0.623 (0.055)           \\
                               & TPR  & \bfseries 0.889 (0.100) & 0.600 (0.516)           & 0.807 (0.084)           & 0.852 (0.089)           \\
                               & PPV  & 0.381 (0.065)           & 0.217 (0.190)           & \bfseries 0.594 (0.039) & 0.499 (0.071)           \\
        \hline
        \multirow[c]{3}{*}{HD} & F1   & 0.549 (0.050)           & 0.252 (0.326)           & 0.581 (0.014)           & \bfseries 0.586 (0.011) \\
                               & TPR  & 0.533 (0.130)           & 0.400 (0.516)           & 0.570 (0.033)           & \bfseries 0.588 (0.030) \\
                               & PPV  & 0.589 (0.034)           & 0.184 (0.238)           & \bfseries 0.594 (0.011) & 0.585 (0.012)           \\
        \hline
        \multirow[c]{3}{*}{PR} & F1   & 0.435 (0.103)           & 0.180 (0.159)           & \bfseries 0.528 (0.089) & 0.507 (0.095)           \\
                               & TPR  & 0.453 (0.098)           & \bfseries 0.600 (0.516) & 0.589 (0.111)           & 0.522 (0.081)           \\
                               & PPV  & 0.498 (0.065)           & 0.106 (0.094)           & 0.501 (0.067)           & \bfseries 0.519 (0.072) \\
        \bottomrule
    \end{tabular}
\end{table}

\subsection{Unlearn Rows}

\Cref{tab:row_fidelity} shows the fidelity of the retrain and three gradient-based unlearning algorithms on the row unlearning task.
The rate of unlearning is 5\% in these experiments which indicates 5\% of the instances are removed from the datasets.
The gradient difference method ranks the best in half of the metrics on four datasets, indicating the effectiveness of gradient difference in unlearning rows.
KL minimization also shows competitive performance in the experiments.
Gradient ascent ranks the worst in all the metrics on all the datasets, as special care is needed in the gradient ascent method to avoid the catastrophic forgetting of knowledge.

Except for the rate of 5\%, we also conduct the experiments with the rate of 10\% and 20\% on the Private dataset.
As shown in \cref{tab:row_fidelity_dynate}, the gradient difference method ranks the best over half of the metrics on all the rates, especially on the F1 score.
Gradient ascent ranks top twice on the TPR metric, however, at the cost of extremely low PPV.
\begin{table}[htbp]
    \centering
    \caption{Fidelity of the algorithms for the row unlearning Private dataset with different unlearning rates.}
    \label{tab:row_fidelity_dynate}
    \renewcommand{\arraystretch}{0.9}
    \begin{tabular}{llrrrr}
        \toprule
        {rate}                   & {loss} & {KL}          & {grad ascent}           & {grad diff}             & {retrain}               \\
        \midrule
        \multirow[c]{3}{*}{0.05} & F1     & 0.435 (0.103) & 0.180 (0.159)           & \bfseries 0.528 (0.089) & 0.507 (0.095)           \\
                                 & TPR    & 0.453 (0.098) & \bfseries 0.600 (0.516) & 0.589 (0.111)           & 0.522 (0.081)           \\
                                 & PPV    & 0.498 (0.065) & 0.106 (0.094)           & 0.501 (0.067)           & \bfseries 0.519 (0.072) \\
        \hline
        \multirow[c]{3}{*}{0.1}  & F1     & 0.417 (0.101) & 0.141 (0.151)           & \bfseries 0.523 (0.089) & 0.516 (0.065)           \\
                                 & TPR    & 0.525 (0.196) & 0.500 (0.527)           & \bfseries 0.575 (0.105) & 0.541 (0.083)           \\
                                 & PPV    & 0.456 (0.141) & 0.082 (0.088)           & 0.506 (0.067)           & \bfseries 0.514 (0.053) \\
        \hline
        \multirow[c]{3}{*}{0.2}  & F1     & 0.274 (0.204) & 0.218 (0.152)           & \bfseries 0.552 (0.051) & 0.499 (0.107)           \\
                                 & TPR    & 0.425 (0.392) & \bfseries 0.700 (0.483) & 0.593 (0.085)           & 0.541 (0.108)           \\
                                 & PPV    & 0.317 (0.296) & 0.129 (0.091)           & \bfseries 0.523 (0.054) & 0.500 (0.067)           \\
        \bottomrule
    \end{tabular}
\end{table}

\Cref{fig:row_DB_0.05_F1} shows the efficiency of the unlearning algorithms on the Diabetes dataset.
Obversely, the gradient difference method is more efficient than the retraining method.
Gradient ascent shows a sharp decrease in the F1 score in the middle stage, which indicates the catastrophic forgetting of knowledge.
KL minimization curve decreases gradually through the training process.

\section{Conclusion}
\label{conclusion}

In summary, this work presents a comprehensive benchmarking study of machine unlearning methods in the federated learning setting for tabular data.
We investigate two dimensions of unlearning with several gradient based methods and fine-tuning.
Extensive experiments are conducted on six datasets to evaluate the fidelity, certifiability, and efficiency of the unlearning algorithms in FL environments.
Generally, finetune and gradient difference methods show competitive performance in two scenarios respectively.
Hopefully, this work provides a foundation for further research in privacy-preserving machine learning.
Especially, the gradient difference method shows promising results in the row unlearning task, which is worth further exploration.

%%%%%%%%%%%%%%%%%%%%%%%%%
% advice from Diederick %
%%%%%%%%%%%%%%%%%%%%%%%%%
% I believe we should enrich the discussion section. Currently, it only focuses on fidelity, certifiability, and efficiency, with efficiency still pending. I would suggest we expand on how our work aligns with previous research. What is our specific contribution? Are we presenting a new method, or are we applying a commonly used method to a different data group? If it’s the latter, we should clarify why this is significant for privacy, security, and regulatory compliance. Alternatively, if our approach is unique in its application (such as in financial services or healthcare), we should highlight this. I think it’s essential to emphasize our innovation more prominently.

\bibliographystyle{splncs04}
\bibliography{ref}

\end{document}